\documentclass[letterpaper]{article}
\usepackage[]{aaai2027}
\usepackage[hyphens]{url}
\usepackage{graphicx}
\usepackage{natbib}
\usepackage{caption}
\usepackage[table]{xcolor}
\definecolor{panelcolor}{RGB}{235,239,242}
\usepackage{amsmath,amssymb}
\usepackage{booktabs}
\usepackage{multirow}
\usepackage{array}
\usepackage{xcolor}
\usepackage{xspace}
\usepackage{listings}
\usepackage{tikz}
\usetikzlibrary{arrows.meta,positioning,fit,calc}

\definecolor{funblue}{HTML}{2F6B9A}
\definecolor{fungreen}{HTML}{5C8A63}
\definecolor{funorange}{HTML}{C8793A}
\definecolor{fungray}{HTML}{F2F4F7}
\newcommand{\method}{\textsc{FunL2O}\xspace}
\newcommand{\code}[1]{\texttt{#1}}
\newcommand{\best}{\textbf}

\title{FunL2O: LLM-Guided Feature Function Design for Learning to Optimize}
\author{
    Bingheng Li\textsuperscript{\rm 1},
    Junyang Cai\textsuperscript{\rm 2},
    Yupeng Zhang\textsuperscript{\rm 3},
    Bistra Dilkina\textsuperscript{\rm 2},\\
    Jayant Kalagnanam\textsuperscript{\rm 4},
    Dzung T. Phan\textsuperscript{\rm 4}
}

\affiliations{
    \textsuperscript{\rm 1} Michigan State University
    \textsuperscript{\rm 2} University of Southern California\\
    \textsuperscript{\rm 3} University of Wisconsin–Madison
    \textsuperscript{\rm 4} IBM Research\\
    libinghe@msu.edu, caijunya@usc.edu, yupeng.zhang@wisc.edu, dilkina@usc.edu,\\ jayant@us.ibm.com,phandu@us.ibm.com
}

\begin{document}
\maketitle

\begin{abstract}
Learning-to-optimize (L2O) methods accelerate repeated optimization by training models to predict solutions, warm starts, branching decisions, or other forms of solver guidance. A critical yet largely overlooked component of these pipelines is the feature function that maps problem instances to inputs for machine learning models. Existing L2O methods typically rely on hand-crafted features, making representation design manual and largely fixed across domains. We introduce FunL2O, the \textbf{first} unified framework for automating feature design through LLM-driven program evolution for L2O. In a FunSearch-style loop, an LLM proposes executable feature functions, while a fixed evaluation process retrains the original L2O model and measures downstream optimization performance. We evaluate FunL2O on linear and quadratic programming tasks involving solution prediction and warm-starting, as well as on mixed-integer optimization tasks using GNN-guided backdoor branching and Predict-and-Search. Across continuous and discrete optimization tasks and four LLMs, the evolved features \textbf{consistently outperform} hand-crafted representations. These results establish LLM-driven feature evolution as a general and effective approach to automating representation design in L2O.

\end{abstract}

\section{Introduction}

Many optimization problems are solved repeatedly with the same mathematical structure but different numerical data. Learning-to-optimize (L2O) methods exploit this regularity by training neural networks on previously solved instances \citep{andrychowicz2016learning,chen2022primer}. Depending on the pipeline, the learned component may predict a solution, initialize an iterative algorithm, select a simplex basis or pivot, or guide mixed-integer search \citep{qian2024power,li2024pdhg,fan2023smart,liu2024pivot,han2023gnn,cai2024multi}. Despite these varied roles, every L2O method begins with the same fundamental design decision: how to represent an optimization instance using features that enable a learning model to train more effectively and capture the relevant problem structure.

The feature function determines what information is directly available to the model and how that information is structured. For linear and mixed-integer programs, the representation may encode objective coefficients, variable bounds and types, graph structure, and properties of the LP relaxation. When a useful relationship is made explicit, the model can exploit it directly; otherwise, it must infer that relationship from the available inputs. Consequently, changing only the input representation can affect both the model’s expressive power and its downstream performance \citep{sun2023feature,chen2025symmetry,cai2025id}.

Despite its importance, the feature function is typically hand-designed and fixed across problem domains before training. Most L2O research then focuses on the model architecture, learning objective, or solver interface while leaving the input representation unchanged. Automating feature design, however, is not simply a matter of feature selection. An L2O feature function is an executable program over structured optimization data: it must preserve a model-specific tensor interface, satisfy semantic and dimensional invariants, and rely only on information available at deployment time. Moreover, its quality can be evaluated only after retraining the learner and measuring downstream performance. The relevant metric is role-dependent: prediction methods prioritize feasibility and objective quality, whereas solver policies seek to reduce iterations, pivots, search effort, or wall-clock time.

To address this problem, we introduce \method{}, the first LLM-guided framework for automatically designing executable feature functions for L2O. The framework defines clear rules for what information a feature function may use at deployment, what output format it must produce, and what conditions it must satisfy. Each valid candidate replaces the original fixed hand-crafted feature function in the L2O pipeline. The model is then retrained, and its downstream optimization performance is used to guide the search toward better feature functions. All other parts of the pipeline—including the model architecture, loss function, labels, training schedule, data, and solver—remain unchanged. This allows us to isolate the effect of the input representation. After the best feature function is selected, it is used as standard preprocessing code and does not require an LLM at deployment.

\begin{figure*}
    \centering
    \includegraphics[width=\linewidth]{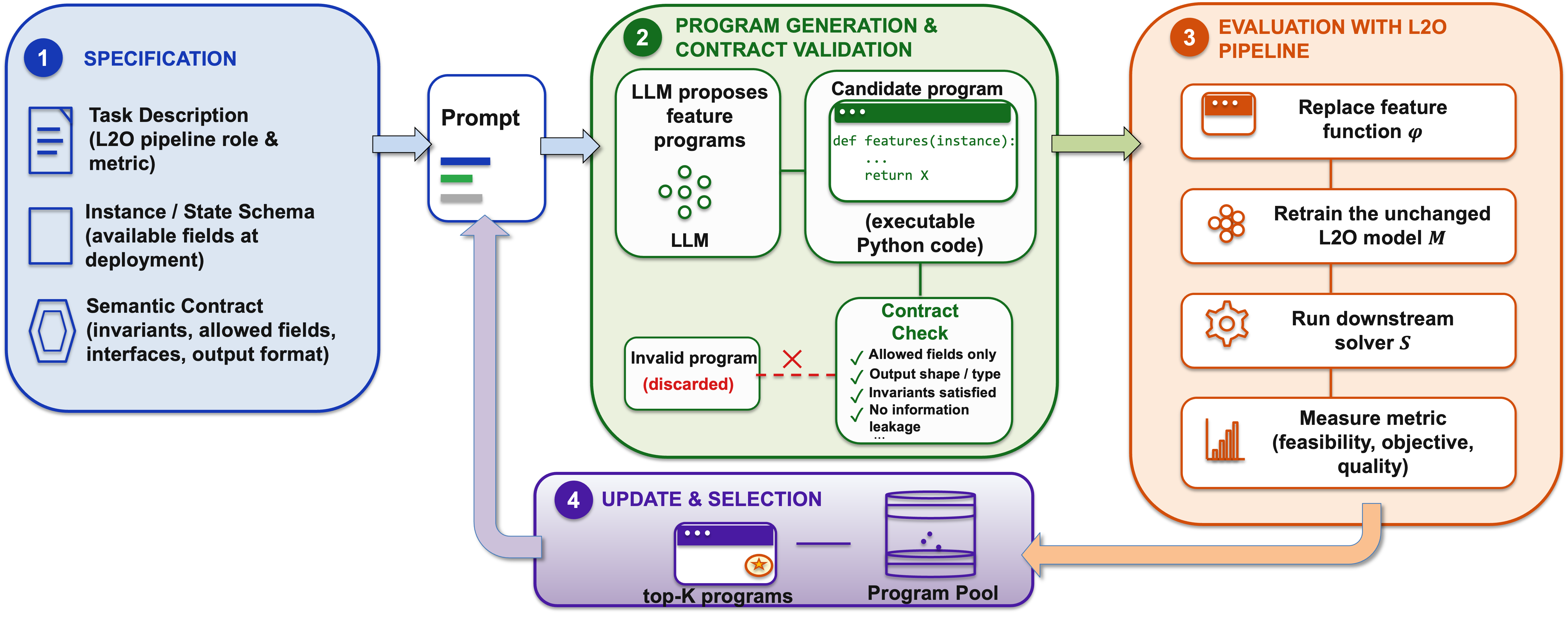}
    \caption{\method{} searches over executable feature functions. Each proposal must satisfy a pipeline-specific semantic contract, after which the original L2O model is retrained and evaluated. Measured optimization outcomes determine the elite programs supplied to the next generation.}
    \label{fig:overview}
\end{figure*}

Our contributions are threefold. First, we formulate feature-function design as a program-search problem within the original L2O pipeline. We also introduce semantic contracts that define valid inputs and outputs for heterogeneous graphs, dense tensors, and solver states. Second, we develop an L2O-in-the-loop search process in which executable feature functions are selected based on measured feasibility and downstream optimization performance, rather than on LLM judgments. Third, we evaluate the framework on eight L2O pipelines covering LP and QP solution prediction, constrained nonlinear prediction, primal--dual warm starting, simplex basis initialization, pivot selection, and MILP search. Across these settings, the evolved features consistently outperform hand-crafted representations, improving solution quality, feasibility, and solver efficiency without changing the learned model or the optimization pipeline.

\section{Background}
\label{sec:background}

\paragraph{Learning to optimize.} 
Learning-to-optimize (L2O) trains a model on related problem instances to predict a solution or guide an optimization algorithm \citep{andrychowicz2016learning,chen2022primer}. Given an instance $x$, an L2O pipeline can be written as \begin{equation} x \xrightarrow{\phi} X \xrightarrow{M} a \xrightarrow{\mathcal{S}} \widehat{y}, \label{eq:l2o-pipeline} \end{equation} where \(\phi\) is the feature function, \(M\) is the learned model, \(a\) is a predicted solution or solver directive, and \(\mathcal{S}\) is the downstream solver. Depending on the learned role, the outcome \(\widehat{y}\) is evaluated by feasibility, objective quality, iterations, pivots, or wall time.

\paragraph{Feature functions.} 
The feature function transforms structured optimization data into tensors consumed by the model. For an LP or MILP, \begin{equation} \min\left\{ c^\top x : Ax \leq b,\; \ell \leq x \leq u,\; x_j \in \mathbb{Z}\ (j\in\mathcal I) \right\}, \label{eq:program} \end{equation} a common representation is a bipartite graph with variable and constraint nodes and an edge for each nonzero coefficient of \(A\) \citep{gasse2019exact}. Features may encode objective coefficients, bounds, variable types, constraint statistics, and LP-relaxation information. These choices determine which numerical and structural relations are directly available to the learner.

\paragraph{Learned solver guidance.}
Some L2O models predict solutions directly, while others guide a classical solver through warm starts, simplex bases, pivot scores, partial assignments, or branching priorities. For solver-facing roles, prediction accuracy is only an intermediate metric: the primary objective is to reduce the work performed by the solver. Accordingly, \method{} evaluates each feature function using the native downstream metric of its host pipeline.

\paragraph{Mixed-integer Prediction.}
We consider two MILP-related prediction tasks. Predict-and-Search predicts binary-variable marginals and searches a trust region around the resulting partial assignment \citep{han2023gnn,huang2024contrastive}. Learned backdoors rank important variables and assign higher branching priorities to the top predictions \citep{cai2024learning}. In both cases, \method{} changes only the feature function, leaving the model, training procedure, and solver fixed.
\section{Methodology}
\label{sec:method}

\method{} takes an existing L2O pipeline \(\mathcal P\) and its handcrafted feature function \(\phi_0\), and returns an evolved feature function as executable code. The search changes only the representation: the model architecture, training procedure, hyperparameters, data, learned role, and downstream solver remain fixed. As shown in Figure~\ref{fig:overview}, an LLM proposes candidate functions, a semantic contract checks their validity, and the original pipeline evaluates valid candidates by retraining the model and measuring the native optimization outcome. The highest-ranked programs guide subsequent proposals.

\subsection{Feature-Function Search}

Consider an L2O pipeline \(\mathcal P\) with a learned model \(M_\theta\), a training procedure, and an optional downstream solver operation \(S_{\mathcal P}\). Let \(x\in\mathcal X_{\mathcal P}\) denote an optimization instance or a solver state. A feature function
\begin{equation}
\phi:
\mathcal X_{\mathcal P}
\rightarrow
\mathcal Z_{\mathcal P}(d_\phi)
\end{equation}
maps \(x\) to a representation of width \(d_\phi\). The model uses this representation to produce a learned output:
\begin{equation}
\begin{split}
z_\phi(x) &= \phi(x),\\
a_\phi(x) &= M_{\theta_\phi^\star}^{(d_\phi)}
\bigl(z_\phi(x)\bigr).
\end{split}
\label{eq:learned_output}
\end{equation}
The learned output is then passed to the remaining optimization pipeline:
\begin{equation}
\widehat y_\phi(x)
=
S_{\mathcal P}
\bigl(x,a_\phi(x)\bigr).
\label{eq:pipeline}
\end{equation}
The output \(a_\phi(x)\) may be a predicted solution, a primal--dual initialization, a basis-status prediction, or a pivot score. When the model directly predicts the final solution, \(S_{\mathcal P}\) is the identity map.

Existing L2O methods provide a handcrafted feature function \(\phi_0\). FunL2O keeps the rest of \(\mathcal P\) fixed and searches for a replacement \(\phi\). A model trained with one representation cannot be reused to evaluate another representation. Therefore, every candidate induces a new training problem:
\begin{equation}
\theta_\phi^\star
=
\operatorname{Train}_{\mathcal P}
\left(
\phi;\mathcal D_{\mathrm{tr}}
\right).
\label{eq:inner_training}
\end{equation}
If \(d_\phi\) differs from the width of the handcrafted representation, the input layer is instantiated at the corresponding width. This is a mechanical interface change: the model template, hidden architecture, loss, optimizer, training schedule, and other hyperparameters are not searched.

After training, the candidate is evaluated on the validation set. We write
\begin{equation}
\mathbf m_{\mathcal P}(\phi)
=
\operatorname{Agg}_{x\in\mathcal D_{\mathrm{val}}}
\mu_{\mathcal P}
\left(
x,\widehat y_\phi(x)
\right)
\label{eq:outcome}
\end{equation}
for its measured outcome, where \(\mu_{\mathcal P}\) is defined by the original pipeline. Depending on the learned role, \(\mathbf m_{\mathcal P}\) may contain feasibility rate, constraint violation, objective gap, iteration count, running time, or pivot count.

These outcomes are compared through a pipeline-specific ranking key
\begin{equation}
\rho_{\mathcal P}(\phi)
=
\left(
b_{\mathcal P}(\phi),
v_{\mathcal P}(\phi),
q_{\mathcal P}(\phi)
\right),
\label{eq:ranking}
\end{equation}
ordered lexicographically with lower values preferred. The first component \(b_{\mathcal P}\) indicates whether the required feasibility condition is violated. The second component \(v_{\mathcal P}\) measures the degree of infeasibility, and the third component \(q_{\mathcal P}\) measures objective quality or solver work. For tasks without a feasibility condition, the first two components are set to zero. The sign of \(q_{\mathcal P}\) is chosen so that lower is always better; for example, it may be objective gap, iteration count, pivot count, or the negative of iteration saving.

The feature-design problem is therefore
\begin{equation}
\phi^\star
\in
\operatorname*{arg\,min}_{\phi\in\Phi_{\mathcal P}}
\rho_{\mathcal P}(\phi).
\label{eq:feature_search}
\end{equation}
The minimization in Equation~\ref{eq:feature_search} follows the lexicographic order. It prevents an infeasible prediction from being preferred only because it has a better objective value. It also allows the same outer problem to represent prediction quality, warm-start effectiveness, and learned solver decisions.

Equation~\ref{eq:feature_search} presents two difficulties. First, the admissible space \(\Phi_{\mathcal P}\) differs across pipelines because their inputs and model interfaces differ. Second, the outer objective is an expensive black box because evaluating one candidate requires model training and may also require a complete solver run. FunL2O addresses the first difficulty through semantic feature contracts and the second through a feedback-driven search over measured outcomes.

\subsection{Semantic Feature Contracts}

A semantic feature contract specifies which feature functions belong to \(\Phi_{\mathcal P}\). For pipeline \(\mathcal P\), we define
\begin{equation}
\mathcal C_{\mathcal P}
=
\left(
\mathcal U_{\mathcal P},
\mathcal Z_{\mathcal P},
\mathcal I_{\mathcal P},
V_{\mathcal P}
\right).
\label{eq:contract}
\end{equation}
Here, \(\mathcal U_{\mathcal P}\) contains the instance fields or solver-state information available at deployment. The output specification \(\mathcal Z_{\mathcal P}\) defines the required return type, orientation, and maximum feature width. The invariants \(\mathcal I_{\mathcal P}\) describe properties that must be preserved, such as determinism, finite outputs, and required handcrafted channels. Finally, \(V_{\mathcal P}\) is an executable validator for these requirements.

\begin{table*}[t] \centering \scriptsize \setlength{\tabcolsep}{3.2pt} \caption{Instantiation of \method{} across the host L2O pipelines. In every case, the feature function is searched while the learned output, training procedure, and downstream solver interface remain fixed.} \label{tab:experiment-overview} \resizebox{\textwidth}{!}{ \begin{tabular}{llllll} \toprule Host method & Training signal & Problem class & Learned output & Primary metric & Secondary metric \\ \midrule IPM-MPNN \citep{qian2024power} & supervised & LP & primal solution & objective gap \(\downarrow\) & feasibility \(\uparrow\) \\ FSNet \citep{nguyen2026fsnet} & self-supervised & QP/QCQP/SOCP & constrained solution & optimality gap \(\downarrow\) & feasibility@\(10^{-4}\) \(\uparrow\) \\ DC3 \citep{donti2021dc3}& self-supervised & constrained QP & corrected solution & optimality gap \(\downarrow\) & feasibility@\(10^{-4}\) \(\uparrow\) \\ PDHG-Net \cite{li2024pdhg} & supervised & PageRank LP & primal--dual warm start & iteration saving \(\uparrow\) & time saving \(\uparrow\) \\ Smart Initial Basis \citep{fan2023smart} & supervised & set-cover LP & initial basis status & iteration saving \(\uparrow\) & node accuracy \(\uparrow\) \\ Learning to Pivot \citep{liu2024pivot} & supervised & set-cover LP & pivot scores & pivot-count gmean \(\downarrow\) & top-1 accuracy \(\uparrow\) \\ Predict-and-Search \cite{han2023gnn} & supervised & MILP & partial assignment & primal gap \(\downarrow\) & primal integral \(\downarrow\) \\ Learned Backdoors \cite{cai2024learning}& supervised & MILP & branching priorities & wall time \(\downarrow\) & win rate \(\uparrow\) \\ \bottomrule \end{tabular}
 } \end{table*}

The contract defines the intended admissible space
\begin{equation}
\Phi_{\mathcal P}
=
\left\{
\phi\in\mathcal H_{\mathrm{safe}}
\;\middle|\;
\substack{
\operatorname{Fields}(\phi)\subseteq\mathcal U_{\mathcal P},\\
\phi(x)\in\mathcal Z_{\mathcal P}(d_\phi),\\
\mathcal I_{\mathcal P}(\phi,x)=1
}
\right\}.
\label{eq:admissible_space}
\end{equation}
Here, \(\mathcal H_{\mathrm{safe}}\) denotes executable functions allowed by the restricted environment. A candidate may combine information already available to the original method, but it cannot access reference solutions, training labels, files, networks, or external solvers.

The output specification remains pipeline-specific. A graph model may require separate variable and constraint features, whereas a constrained predictor may expect a dense tensor and a pivot model may expect one feature vector per solver-state node. This allows FunL2O to share one search procedure without forcing different L2O methods into a common representation. Table~\ref{tab:experiment-overview} summarizes how the semantic contract and native evaluation criterion are instantiated across the host pipelines. Although the feature interface and learned role vary, the proposal, validation, retraining, and selection procedure remains unchanged. 

When the original representation must be retained, the contract requires the candidate to preserve its channels. In the common append-only case, the candidate has the form
\begin{equation}
\phi(x)
=
\operatorname{Concat}
\left(
\phi_0(x),\psi(x)
\right),
\label{eq:seed_preservation}
\end{equation}
where \(\psi\) contains the newly proposed features. For structured outputs, the same requirement is applied separately to the relevant variable, constraint, edge, or global channels.

In implementation, \(V_{\mathcal P}\) first applies a static guard to the candidate source and then executes it on a small probe instance. The validator checks the function signature, output structure, feature dimensions, numerical finiteness, and required seed channels. A failed check returns the violated condition to the proposal model. Passing \(V_{\mathcal P}\) establishes interface validity only; whether a function is useful is determined by retraining and evaluating it according to Equations~\ref{eq:inner_training}--\ref{eq:ranking}.


\subsection{Standardized Prompt Construction}

The semantic contract determines whether a feature function can be used by the original pipeline. To propose a useful function, the LLM must also know how candidates are evaluated and which relations have already been explored. FunL2O organizes this information through one prompt template shared by all pipelines.

At generation \(g\), the prompt is constructed as
\begin{equation}
p_g
=
\mathcal T
\left(
\mathcal R,
\mathcal C_{\mathcal P},
\mathcal G,
\mathcal E_g,
\mathcal F_g
\right).
\label{eq:prompt}
\end{equation}
The shared instructions \(\mathcal R\) state that only the feature function may change and that the returned function must be deterministic and executable. The contract \(\mathcal C_{\mathcal P}\) supplies the available inputs, function signature, output structure, and pipeline-native evaluation rule.

The shared vocabulary \(\mathcal G\) describes broad relations that may be useful in optimization representations, including raw problem quantities, scale, residual, curvature, bound, cone, and graph statistics. These groups guide the proposal model but do not define a fixed library of allowed features. The generation-specific context \(\mathcal E_g\) contains the source code, feature width, and validation outcome of the current elite functions. The feedback \(\mathcal F_g\) contains a contract error or a summary of whether the preceding generation improved the current best.

Given this prompt, the proposal model \(Q_\omega\) generates a population of candidate feature functions:
\begin{equation}
\widetilde\Phi_g
\sim
Q_\omega
\left(
\,\cdot\mid p_g
\right).
\label{eq:proposal}
\end{equation}
The model is asked to identify a useful relation that the current representation may not expose and to implement it as one feature function. It receives aggregate validation outcomes and candidate source code, but it does not receive validation instances, target solutions, or test results. The LLM therefore proposes feature computations; it does not predict their performance.

\subsection{L2O-in-the-Loop Search}

The search begins by evaluating the handcrafted function \(\phi_0\). Its source code, feature width, measured outcome, and ranking key initialize the search memory:
\begin{equation}
\mathcal M_0
=
\left\{
\left(
\phi_0,
d_{\phi_0},
\mathbf m_{\mathcal P}(\phi_0),
\rho_{\mathcal P}(\phi_0)
\right)
\right\}.
\label{eq:initial_memory}
\end{equation}
The handcrafted function therefore serves both as the performance baseline and as the first valid example of the semantic contract.

\begin{table*}[t] 
\centering 
\setlength{\tabcolsep}{4.0pt} 
\caption{Results across continuous-optimization L2O pipelines. Prediction gaps and solver costs are lower-is-better; saving metrics and accuracy are higher-is-better. Prediction results show clean-test medians over four paired searches. Feasibility is reported as handcrafted to \method{}. Positive IR always denotes improvement.} 
\label{tab:continuous-main} 
\begin{tabular}{lllrrc}
\toprule
Host & Task or metric & Handcrafted & \method{} & Improvement & Feasibility (H \(\to\) F) \\
\midrule

\rowcolor{panelcolor}
\multicolumn{6}{l}{\textit{Panel A: LP solution prediction}} \\
\midrule

\multirow{5}{*}{IPM-MPNN}
& Independent set objective gap & 0.00365 & \textbf{0.00306} & \textbf{16.4\%} & 0.914 \(\to\) 0.920 \\
& Set cover objective gap & 0.01207 & \textbf{0.00611} & \textbf{49.4\%} & 0.957 \(\to\) 0.942 \\
& Combinatorial auction objective gap & 0.02217 & \textbf{0.01660} & \textbf{25.1\%} & 0.874 \(\to\) 0.883 \\
& Facility location objective gap & 0.02303 & \textbf{0.01813} & \textbf{21.3\%} & 0.988 \(\to\) 0.941 \\
& GNNLP objective gap & 0.00499 & \textbf{0.00383} & \textbf{23.2\%} & 0.979 \(\to\) 0.979 \\
\midrule

\rowcolor{panelcolor}
\multicolumn{6}{l}{\textit{Panel B: Constrained solution prediction}} \\
\midrule

\multirow{9}{*}{FSNet}
& Convex QP gap & 0.685 & \textbf{0.363} & \textbf{47.0\%} & 0.938 \(\to\) 1.000 \\
& Convex QCQP gap & 14.09 & \textbf{8.61} & \textbf{38.9\%} & 1.000 \(\to\) 1.000 \\
& Convex SOCP gap & 42.33 & \textbf{22.52} & \textbf{46.8\%} & 1.000 \(\to\) 1.000 \\
& Nonconvex QP gap & 5.46 & 5.46 & 0.0\% & 1.000 \(\to\) 1.000 \\
& Nonconvex QCQP gap & 26.42 & \textbf{19.26} & \textbf{27.1\%} & 1.000 \(\to\) 1.000 \\
& Nonconvex SOCP gap & 933.85 & \textbf{141.28} & \textbf{84.9\%} & 0.250 \(\to\) 1.000 \\
& Nonsmooth nonconvex QP gap & 5.20 & 5.20 & 0.0\% & 1.000 \(\to\) 1.000 \\
& Nonsmooth nonconvex QCQP gap & 22.69 & 22.69 & 0.0\% & 1.000 \(\to\) 1.000 \\
& Nonsmooth nonconvex SOCP gap & 332.33 & \textbf{159.38} & \textbf{52.0\%} & 0.750 \(\to\) 1.000 \\
\cmidrule(lr){1-6}

\multirow{3}{*}{DC3}
& Convex QP gap & 13.01 & \textbf{12.65} & \textbf{2.8\%} & 0.986 \(\to\) 0.986 \\
& Nonconvex QP gap & 15.62 & \textbf{15.31} & \textbf{2.0\%} & 0.848 \(\to\) 0.849 \\
& Nonsmooth nonconvex QP gap & 15.96 & \textbf{15.28} & \textbf{4.3\%} & 0.855 \(\to\) 0.861 \\
\midrule

\rowcolor{panelcolor}
\multicolumn{6}{l}{\textit{Panel C: Solver-facing learned roles}} \\
\midrule

\multirow{2}{*}{PDHG-Net}
& Iteration saving \(\uparrow\) & 42.44\% & \textbf{51.05\%} & \textbf{20.3\%} & -- \\
& Time saving \(\uparrow\) & 34.03\% & \textbf{39.28\%} & \textbf{15.4\%} & -- \\
\cmidrule(lr){1-6}

Smart Initial Basis
& Iteration saving \(\uparrow\) & 59.28\% & \textbf{60.24\%} & \textbf{1.62\%} & -- \\
\cmidrule(lr){1-6}

\multirow{2}{*}{Learning to Pivot}
& Pivot-count geometric mean \(\downarrow\) & 1162.86 & \textbf{857.78} & \textbf{26.2\%} & -- \\
& Top-1 accuracy \(\uparrow\) & 0.5264 & \textbf{0.5313} & \textbf{0.9\%} & -- \\
\bottomrule
\end{tabular}
\end{table*}

At generation \(g\), FunL2O constructs \(p_g\) from the contract and the current memory, and samples the proposed population \(\widetilde\Phi_g\). Candidates that pass the validator form
\begin{equation}
\Phi_g
=
\left\{
\phi\in\widetilde\Phi_g
:
V_{\mathcal P}(\phi)=1
\right\}.
\label{eq:valid_population}
\end{equation}
If validation fails, the violated condition is appended to the prompt and the proposal model is asked to repair the function. A function is submitted for training only after it passes the contract.

Every \(\phi\in\Phi_g\) replaces \(\phi_0\) in a fresh copy of the original pipeline. The corresponding model is trained using the original data, loss, optimizer, and schedule, and its outcome is measured by \(\mu_{\mathcal P}\). Candidates within a generation are independent given the current memory and can therefore be trained and evaluated in parallel.

After evaluation, the search memory is updated as
\begin{equation}
\begin{split}
\mathcal M_{g+1}
=
\mathcal M_g
\cup
\bigl\{
(&\phi,d_\phi,\mathbf m_{\mathcal P}(\phi),\\
 &\rho_{\mathcal P}(\phi))
:
\phi\in\Phi_g
\bigr\}.
\end{split}
\label{eq:memory_update}
\end{equation}
The elite functions supplied to the next prompt are selected by
\begin{equation}
\mathcal E_{g+1}
=
\operatorname{TopK}_{\rho_{\mathcal P}}
\left(
\mathcal M_{g+1}
\right),
\label{eq:elite_update}
\end{equation}
where \(\operatorname{TopK}_{\rho_{\mathcal P}}\) returns the \(K\) entries with the lowest pipeline-specific ranking keys. The same proposal, validation, evaluation, and memory update are used for every L2O pipeline. Only the contract \(\mathcal C_{\mathcal P}\), outcome measure \(\mu_{\mathcal P}\), and ranking key \(\rho_{\mathcal P}\) change. After the final generation, FunL2O returns the lowest-ranked valid feature function in the memory. The selected function is ordinary executable preprocessing code. Neither the proposal model nor the search memory is required during deployment.

\section{Experiments}
\label{sec:experiments}


\subsection{Experimental Setup}

Table~\ref{tab:experiment-overview} summarizes the eight host pipelines. The continuous-optimization study includes IPM-MPNN for LP solution prediction \citep{qian2024power}; FSNet and DC3 for constrained prediction \citep{donti2021dc3,nguyen2026fsnet}; PDHG-Net for PageRank LP warm starts \citep{li2024pdhg}; and Smart Initial Basis and Learning to Pivot for set-cover simplex decisions \citep{fan2023smart,liu2024pivot}. The MILP study includes Predict-and-Search and learned backdoor branching on CA, MIS, and MVC instances \citep{han2023gnn,cai2024learning}. In every experiment, \method{} modifies only the feature function.

For all pipelines, candidate feature functions are selected on validation instances and evaluated on a disjoint test set, with no overlap among the training, validation, and test instances. Each search consists of eight iterations with six proposal slots. We repeat every experiment with three different random seeds and report the mean. For the continuous-optimization experiments, we run \method{} with four frontier LLMs: Claude-Opus-4.8 \citep{anthropic2026claudeopus48}, GPT-5.5 \citep{openai2026gpt55}, Gemini-3.1-Pro Preview \citep{google2026gemini31pro}, and DeepSeek-Math-V2 \citep{shao2025deepseekmath}. For the MILP experiments, we use Claude-Opus-4.8 and GPT-5.5. Predict-and-Search runs CPLEX \citep{manual1987ibm} with a 1000-second time limit, whereas the backdoor-branching experiments use a 600-second CPLEX time limit.

\begin{table*}[t]
\centering
\setlength{\tabcolsep}{5.0pt}
\caption{Results for mixed-integer search. For Predict-and-Search, final primal gap and primal integral are lower-is-better. For learned backdoors, mean wall time is lower-is-better and win rate is higher-is-better. \best{Bold} indicates the better evolved feature function in each row.}
\label{tab:mip-main}
\begin{tabular}{llrrrr}
\toprule
Domain & Metric & Default Solver & Handcrafted & \method{}-Claude & \method{}-GPT \\
\midrule

\rowcolor{panelcolor}
\multicolumn{6}{l}{\textit{Panel A: Predict-and-Search}} \\
\midrule

CA & final primal gap (\%) & 0.841 & 0.423 & \best{0.350} & 1.023 \\
& primal integral & 11.23 & 10.61 & \best{9.00} & 14.18 \\
\addlinespace
MIS & final primal gap (\%) & 1.317 & 0.699 & 0.213 & \best{0.095} \\
& primal integral & 25.64 & 20.89 & \best{12.78} & 14.27 \\
\addlinespace
MVC & final primal gap (\%) & 0.032 & 0.018 & \best{0.013} & 0.015 \\
& primal integral & 20.08 & 13.85 & 7.40 & \best{3.69} \\

\midrule
\rowcolor{panelcolor}
\multicolumn{6}{l}{\textit{Panel B: Learned backdoor branching}} \\
\midrule

CA & mean wall time (s) & 167.2 & 154.6 & 148.4 & \best{147.6} \\
& win rate (\%) & 8 & 20 & 31 & \best{41} \\
\addlinespace
MIS & mean wall time (s) & 102.6 & 59.1 & 50.2 & \best{46.0} \\
& win rate (\%) & 0 & 21 & 30 & \best{49} \\
\addlinespace
MVC & mean wall time (s) & 24.7 & 31.6 & \best{18.9} & 20.4 \\
& win rate (\%) & 13 & 4 & 37 & \best{46} \\

\bottomrule
\end{tabular}
\end{table*}

\paragraph{Metrics.} 
Solution-prediction hosts retain their original objective gap and feasibility criteria. IPM-MPNN reports the relative objective gap, while FSNet and DC3 report the gap in percent and measure feasibility using a constraint-violation threshold of \(10^{-4}\).

For solver-facing pipelines, iteration saving is $100\frac{C-H}{C}, $ where \(C\) and \(H\) are the cold-start and learned-start iteration counts; time saving is defined analogously. Learning to Pivot reports the geometric mean pivot count. For a lower-is-better metric with handcrafted result \(B\) and \method{} result \(F\), the improvement rate is $ \operatorname{IR} = 100\frac{B-F}{|B|}. $ For higher-is-better metrics, the numerator is \(F-B\), so positive IR always denotes improvement.

For P\&S, final primal gap and primal integral are lower-is-better, with the primal integral additionally rewarding methods that find strong incumbents earlier. Learned backdoors are evaluated by mean wall time. We also report the percentage of instances on which each method is the fastest.

\subsection{Results Across L2O Pipelines} 
\label{sec:main-results}

\paragraph{Continuous optimization.} 
Table~\ref{tab:continuous-main} summarizes all continuous-optimization results. On IPM-MPNN, \method{} reduces the median objective gap on all five LP distributions. The reductions range from \(16.4\%\) on independent set to \(49.4\%\) on set cover, demonstrating consistent gains across structurally different LP families.

The improvements extend to self-supervised constrained prediction. On the nine 1000-variable FSNet tasks, \method{} improves six displayed medians and matches the remaining three. The largest gains occur on SOCP-constrained problems, where the gap decreases by \(46.8\%\), \(84.9\%\), and \(52.0\%\) across the convex, nonconvex, and nonsmooth nonconvex objectives. On nonconvex SOCP, the gap falls from \(933.85\%\) to \(141.28\%\), while feasibility increases from \(0.25\) to \(1.00\). DC3 also improves across all three QP variants, with reductions of \(2.0\%\)--\(4.3\%\).

The evolved features also improve solver-facing learned roles. For PDHG-Net, iteration saving increases from \(42.44\%\) to \(51.05\%\), and time saving increases from \(34.03\%\) to \(39.28\%\). Learning to Pivot reduces the geometric mean pivot count from \(1162.86\) to \(857.78\), a \(26.2\%\) improvement. Its top-1 accuracy changes only modestly, indicating that downstream solver progress can improve substantially even when local prediction accuracy changes little. Smart Initial Basis also increases iteration saving, from \(59.28\%\) to \(60.24\%\).

\paragraph{Mixed-integer Prediction.} 
Table~\ref{tab:mip-main} evaluates evolved feature functions for two learned MILP search strategies. For Predict-and-Search, the best \method{} variant outperforms the handcrafted features on both metrics across all three domains. The final primal gap decreases from \(0.423\%\) to \(0.350\%\) on CA, from \(0.699\%\) to \(0.095\%\) on MIS, and from \(0.018\%\) to \(0.013\%\) on MVC. The corresponding primal integrals decrease from \(10.61\) to \(9.00\), from \(20.89\) to \(12.78\), and from \(13.85\) to \(3.69\), respectively. These results indicate that the evolved representations help Predict-and-Search find stronger incumbents earlier. 

For learned backdoor branching, both \method{}-Claude and \method{}-GPT reduce mean wall time relative to the handcrafted features on all three domains. The best evolved functions achieve mean wall times of \(147.6\) seconds on CA, \(46.0\) seconds on MIS, and \(18.9\) seconds on MVC. Relative to the default solver, these results correspond to reductions of \(11.7\%\), \(55.2\%\), and \(23.5\%\), respectively. The evolved functions also achieve the highest win rates on every domain, reaching \(41\%\) on CA, \(49\%\) on MIS, and \(46\%\) on MVC. Together, these results show that feature evolution extends beyond continuous optimization to learned partial assignments and branching priorities for MILP search.

\subsection{Analysis of the Evolved Features} \label{sec:analysis}

\paragraph{Semantic structure versus feature width.} We compare evolved functions with matched-width random projections wherever the host interface supports an append-only control. On PDHG-Net, iteration saving rises from \(35.59\%\) with handcrafted features to \(45.19\%\) with random features and \(47.99\%\) with \method{}. Evolved functions also outperform matched-width random controls on four of the five IPM-MPNN distributions. These comparisons indicate that the gains often reflect optimization-relevant structure beyond the effect of increasing the input dimension. 

\paragraph{Inspecting the evolved programs.} The returned programs expose recognizable optimization relationships. For set cover, they normalize variable costs by column coverage and constraint signals by row coverage. For PageRank LPs, they compare row and column mass and degree, exposing asymmetry in the directed graph. For nonlinear problems, they construct multiscale representations using magnitude, sign, logarithmic, squared, normalized, and local-difference transformations. The MILP programs combine variable types and objective coefficients with LP-relaxation values, fractionality, reduced costs, row and column statistics, and normalized constraint signals. Because the selected functions are ordinary preprocessing code, these transformations can be inspected directly and deployed without an LLM. 

\paragraph{Proposal-model dependence.}
Across the continuous experiments shared by all four LLMs, Claude-Opus-4.8 , GPT-5.5, Gemini-3.1-Pro Preview, and DeepSeek-Math-V2 improve 15, 17, 17, and 6 cells, respectively. In the MILP experiment, Claude-Opus-4.8 and GPT-5.5 both produce feature functions that reduce backdoor-branching wall time across CA, MIS, and MVC, while the best proposal-model variant also improves both Predict-and-Search metrics on all three domains. These results show that \method{} is not tied to a single LLM provider, although stronger proposal models generate successful feature functions more reliably.

\section{Related Work}
\paragraph{Learning to optimize.}
Learning-to-optimize replaces or augments part of an optimization procedure with a model trained on related instances \citep{andrychowicz2016learning,chen2022primer}. Prior work learns update rules, predicts solutions, or guides classical solvers through warm starts, basis selection, pivoting, partial assignments, and branching decisions \citep{wichrowska2017learned,chen2020stronger,harrison2022closer,yang2023ml2o,donti2021dc3,nguyen2026fsnet,qian2024power,sambharya2024warm,li2024pdhg,fan2023smart,liu2024pivot,han2023gnn,cai2024learning}. These methods typically keep the input representation fixed. In contrast, \method{} searches over executable feature functions while leaving the model, training procedure, prediction target, and solver interface unchanged.

\paragraph{Representations for optimization.}
Optimization problems are often represented as variable--constraint bipartite graphs \citep{gasse2019exact}. Their node and edge features determine what information is available to the model and can limit expressiveness even when the architecture is fixed \citep{chen2025symmetry}. Prior work has proposed specialized or manually designed features for particular problem classes and applications \citep{qian2024power,cai2025id,cai2025neuro}. \method{} instead automates representation design by evaluating executable feature functions through the complete L2O pipeline.

\paragraph{LLM-guided program and feature search.}
FunSearch and related systems use LLMs to generate executable programs that are selected by an external evaluator \citep{romeraparedes2024funsearch,liu2024eoh,ye2024reevo,vanstein2025llamea,novikov2025alphaevolve}. LLMs have also been used to construct features for tabular prediction \citep{hollmann2023caafe,abhyankar2026llmfe}. \method{} applies this idea to structured optimization data and solver states: candidates must satisfy a pipeline-specific semantic contract and are evaluated only after retraining the unchanged L2O model and measuring downstream optimization performance.
\section{Discussion} 
\method{} automates a previously fixed component of the L2O pipeline: the feature function. Whereas existing methods learn solver actions from hand-designed representations, \method{} optimizes the representation itself using downstream optimization performance. This enables a unified framework to improve pipelines with different model architectures, training objectives, and learned outputs.

The evolved programs are especially effective when handcrafted features omit useful structural relationships. They often combine problem-specific quantities, such as cost and coverage, PageRank imbalance, nonlinear transformations, or LP-relaxation statistics. Because the output is executable code, each evolved program serves both as a deployable feature function and as an interpretable hypothesis about what information the original representation lacked. We intentionally use a simple evolutionary loop so that performance gains can be attributed primarily to feature quality rather than to search-engineering choices. More sophisticated selection, mutation, and crossover mechanisms are complementary to our contribution and could be incorporated in future work.

A natural concern is that program search may overfit the finite set of instances used during evolution. To reduce this risk, all candidate selection is performed on validation instances, while final performance is reported only on disjoint test sets. The consistent test-set gains suggest that the evolved features generalize beyond the instances used for candidate ranking. The main limitation is the offline search cost, since each valid candidate requires retraining the host model and evaluating downstream performance. However, the LLM is not needed after the search, so deployment cost remains unchanged. In repeated-optimization settings, this offline cost can be amortized over many future instances \citep{chen2022primer}.
\section{Conclusion} 

L2O methods learn how to act on optimization problems, but typically inherit a handcrafted representation of those problems. \method{} makes this representation searchable. Within a semantic contract, an LLM proposes executable feature functions whose value is determined by retraining the unchanged L2O pipeline and measuring downstream optimization performance. Across LP, QP, constrained nonlinear, and mixed-integer optimization, evolved features improve solution prediction, warm starting, simplex decisions, and MILP search without changing the underlying models or solvers. The selected programs require no LLM at deployment and expose interpretable relations that the original representations omitted. These results establish feature-function design as a distinct and automatable component of learning to optimize.

\section{Acknowledgment}
This paper reports on research conducted while Bingheng Li, Junyang Cai, and Yupeng Zhang interned at IBM Research, Yorktown Heights, New York, US.

\bibliography{references}
\newpage

\appendix
\onecolumn
\section{Experimental and Reproducibility Details}
\label{app:overview}

This appendix documents the final experimental protocol used throughout the paper. The same principle is applied to every host pipeline: \method{} replaces the executable feature function while retaining the host model, learning objective, training schedule, data generator or instance lists, learned output, and downstream solver interface. Candidate quality is determined by the host evaluator rather than by the proposal model, and the returned feature function is used as ordinary deterministic preprocessing code at deployment. Unless stated otherwise, every reported handcrafted--selected comparison is the arithmetic mean of three independent, seed-paired repetitions of the corresponding host training and evaluation protocol.

\paragraph{Computational cost.}
Each \method{} search evaluates 48 candidate feature functions. The average LLM API cost is approximately \$4 per proposal, resulting in a total LLM cost of 48×\$4=\$192 per search. Retraining each candidate takes approximately 15 GPU-minutes, for a total of 48×15=720 GPU-minutes, or 12 GPU-hours. Downstream CPU-based evaluation takes approximately 1000 seconds per candidate, resulting in 48,000 CPU-seconds, or 13.3 CPU-hours, per search. With one GPU and one CPU worker and no parallelization, a complete search requires approximately 25.3 hours of computation, excluding LLM response latency and pipeline overhead. These costs scale linearly with the number of pipeline--task--proposal-model combinations, although candidate evaluations can be parallelized in practice. Once the search is complete, the selected feature function incurs no additional LLM cost and only standard preprocessing overhead at deployment.

\subsection{Host Pipelines and Training Configurations}

Table~\ref{tab:app-config} summarizes the eight pipeline instantiations. The continuous study covers solution prediction, primal--dual warm starts, and simplex decisions. The mixed-integer study applies the same feature-search interface to Predict-and-Search and learned backdoor branching.

\begin{table*}[h]
\centering
\scriptsize
\caption{Host configurations. Only the feature function is searched. The learned output, network template, training objective, and solver interface remain those of the corresponding host implementation.}
\label{tab:app-config}
\resizebox{\textwidth}{!}{%
\begin{tabular}{lllllll}
\toprule
Host & Tasks & Searched function & Training configuration & Learned role & Primary metric & Secondary metric \\
\midrule
IPM-MPNN & five LP distributions & \code{compute\_features} & 80 epochs; hidden size 128 & primal solution & objective gap $\downarrow$ & feasibility $\uparrow$ \\
FSNet & nine main QP/QCQP/SOCP tasks (+ three auxiliary size checks) & \code{compute\_fsnet\_input\_features} & 40 epochs; hidden size 1024 & constrained solution & optimality gap $\downarrow$ & feasibility $\uparrow$ \\
DC3 & three constrained QPs & \code{compute\_dc3\_extra} & 200 epochs & corrected solution & optimality gap $\downarrow$ & feasibility $\uparrow$ \\
PDHG-Net & PageRank LP & \code{compute\_pdhg\_features} & 300 epochs & primal--dual warm start & iteration saving $\uparrow$ & time saving $\uparrow$ \\
Smart Initial Basis & set-cover LP & \code{compute\_sib\_extra} & 200 epochs & initial basis status & iteration saving $\uparrow$ & node accuracy $\uparrow$ \\
Learning to Pivot & set-cover LP & \code{compute\_ltp\_extra} & 20 epochs & pivot scores & pivot-count gmean $\downarrow$ & top-1 accuracy $\uparrow$ \\
Predict-and-Search & CA/MIS/MVC & \code{compute\_variable\_features} & 300 epochs; $k=8$ & partial assignment & primal gap $\downarrow$ & primal integral $\downarrow$ \\
Learned Backdoors & CA/MIS/MVC & \code{compute\_variable\_features} & 300 epochs; $k=8$ & branching priorities & wall time $\downarrow$ & win rate $\uparrow$ \\
\bottomrule
\end{tabular}}
\end{table*}

\subsection{Data and Evaluation Splits}

Every host follows the same separation principle. A candidate feature program is trained using the host training split and ranked exclusively by its validation outcome. After the search terminates, the selected program is frozen and evaluated on a disjoint test split that is never exposed to the search loop or the proposal model. Training, validation, and test instances are disjoint. Different proposal-model runs use the same split definition, so their paired outcomes are directly comparable. Within each reported comparison, the handcrafted feature and the validation-selected feature are evaluated with the same three training seeds, initialization seeds, data-order seeds, instance splits, and solver settings. This pairing removes training-seed variation from the method difference before the three paired outcomes are averaged.

\begin{table*}[t]
\centering
\scriptsize
\caption{Data organization used by the host evaluators. ``Selection'' denotes the data used to rank candidate feature programs; ``reporting'' denotes the data used for the displayed outcome.}
\label{tab:app-splits}
\resizebox{\textwidth}{!}{%
\begin{tabular}{llll}
\toprule
Host & Training data & Selection data & Reporting data \\
\midrule
IPM-MPNN & 70\% of each LP distribution & 15\% validation holdout & disjoint 15\% test holdout \\
FSNet & official training split & official validation split & official disjoint test split \\
DC3 & host training split & held-out validation split & disjoint test split \\
PDHG-Net & host training records & disjoint validation records & 20 held-out test records \\
Smart Initial Basis & 60\% of 1,400 set-cover instances & 15\% validation split & disjoint 25\% test split \\
Learning to Pivot & 50,000 training pivot states & 10,000 validation states & 50 held-out test instances \\
Predict-and-Search & 160 domain-specific MILP training list & 40 disjoint domain validation list & disjoint 100-instance test list \\
Learned Backdoors & 160 domain-specific MILP training list & 40 disjoint domain validation list & disjoint 100-instance test list \\
\bottomrule
\end{tabular}}
\end{table*}

\section{Search Procedure}
\label{app:search}

\subsection{Budget and Elitism}

The handcrafted feature function is evaluated once before program generation and remains in the search memory as the fixed baseline. Each search then runs eight proposal generations with six proposal slots per generation, giving a budget of 48 LLM-proposed candidate programs for every pipeline, task, and proposal model. The two highest-ranked programs form the elite context for the next generation. The same eight-generation, six-proposal, two-elite protocol is used for the continuous and mixed-integer hosts.

Each proposal is compiled and checked against the host contract. A rejected proposal receives the localized validation error and may be repaired up to three times. Valid candidates within a generation are independent conditioned on the current elite memory and are therefore trained and evaluated in parallel.

\noindent\begin{minipage}{\linewidth}
\begin{lstlisting}[basicstyle=\ttfamily\fontsize{6.5}{7.0}\selectfont,
                   frame=single,breaklines=true,columns=fullflexible,
                   showstringspaces=false,aboveskip=3pt,belowskip=3pt]
memory = [train_and_validate(handcrafted_seed)]
for generation in range(8):
    elites = top_two_by_validation(memory)
    context = contract + elites + validation_scores(elites)
    for program in propose(context, count=6):
        program = validate_or_repair(program, max_repairs=3)
        if program is valid:
            val = train_and_validate(program)
            memory.append((program, val))
selected = best_by_validation(memory)
test_result = evaluate_once_on_disjoint_test(selected)
return selected, test_result
\end{lstlisting}
\end{minipage}

The proposal model receives the semantic contract, complete source of the elite feature programs, and aggregate validation outcomes. It never receives validation instances, targets, or test outcomes, and it does not assign fitness. The final-test evaluator is called only after validation-based program selection. After search, neither the proposal model nor the evolutionary memory is needed for inference.

\subsection{Ranking}

For prediction hosts with feasibility requirements, ranking is lexicographic: feasibility acceptance, degree of violation, and then objective quality. This prevents an objective-only improvement from displacing a feasible program. For solver-facing hosts, the native host metric determines the ordering. The handcrafted seed remains in the memory throughout search and is returned whenever no proposal ranks above it.

\section{Semantic Feature Contracts}
\label{app:contracts}

A semantic feature contract defines the callable interface, permitted deployment-time information, tensor orientation, feature-width limit, and required handcrafted prefix. Table~\ref{tab:contract-details} lists the principal interfaces. The contracts permit deterministic transformations of information already exposed by the host and exclude training labels, reference solutions, evaluation outcomes, file or network access, and additional solver calls inside the feature function.

\begin{table*}[t]
\centering
\scriptsize
\caption{Feature-function interfaces and output requirements. $m$ and $n$ denote the numbers of constraints and variables, respectively; $B$ denotes batch size.}
\label{tab:contract-details}
\resizebox{\textwidth}{!}{%
\begin{tabular}{llll}
\toprule
Host & Function signature & Required output & Preserved interface \\
\midrule
IPM-MPNN & \code{compute\_features(A,b,c,sense,lb,ub)} & variable, constraint, and global tensors & first two channels of each node type \\
FSNet & \code{compute\_fsnet\_input\_features(problem,state)} & $B\times F$ dense tensor & raw state tensor as leading channels \\
DC3 & \code{compute\_dc3\_extra(instance)} & $B\times K$, $K\leq64$ & appended to the host input \\
PDHG-Net & \code{compute\_pdhg\_features(A,b,c,sense,lb,ub)} & variable and constraint tensors & 3 variable and 4 constraint seed channels \\
Smart Initial Basis & \code{compute\_sib\_extra(lp)} & $n\times K$ and $m\times K$, $K\leq32$ & appended to locked seed channels \\
Learning to Pivot & \code{compute\_ltp\_extra(base)} & $(n+m)\times K$, $K\leq32$ & appended to the 11-dimensional host input \\
MILP hosts & \code{compute\_variable\_features(model,relaxation)} & per-variable $n\times F$ tensor & host graph and solver interface unchanged \\
\bottomrule
\end{tabular}}
\end{table*}

\paragraph{Static validation.}
Candidate source is restricted to the requested function and a safe execution environment. Imports are sanitized, and the validator prevents external I/O, dynamic code execution, mutable global state, and unavailable information sources.

\paragraph{Dynamic validation.}
A host-specific probe checks the function signature, return structure, row or batch alignment, feature width, numerical finiteness, and required seed channels. Passing the probe establishes interface validity; utility is determined only by retraining and downstream evaluation.

\section{Prompt and Proposal-Model Configuration}
\label{app:prompt}

The same run-independent prompt structure is instantiated with a host-specific function signature, metric, and contract. The complete prompt files are included in the accompanying source artifact. The central template is:

\noindent\begin{minipage}{\linewidth}
\begin{lstlisting}[basicstyle=\ttfamily\fontsize{6.35}{6.9}\selectfont,
                   frame=single,breaklines=true,columns=fullflexible,
                   showstringspaces=false,aboveskip=3pt,belowskip=3pt]
You design input features for a FIXED learning-to-optimize
pipeline. The model, data, training procedure, solver, and
seed baseline are fixed. Change only {function_name}.

Use only fields listed in the semantic feature contract.
Return deterministic, finite features with the required
shape and preserved seed channels. Do not access targets,
reference solutions, files, networks, or additional solvers.

Metric and direction: {host_metric}
Function signature and typed output: {contract}
Elite programs and measured outcomes: {elite_memory}

Return exactly one Python code block defining the function.
\end{lstlisting}
\end{minipage}

The recorded API model identifiers are \code{claude-opus-4-8}, \code{azure/gpt-5.5}, \code{gcp/gemini-3.1-pro-preview}, and \code{rits/deepseek-ai/DeepSeek-Math-V2}. The continuous study uses all four proposal models; the displayed MILP study reports Claude and GPT-5.5 runs separately. The maximum completion length is 16,000 tokens. Temperature and top-$p$ are not explicitly overridden in the search client and therefore follow the provider defaults associated with the recorded endpoints.

\section{Metrics and Aggregation}
\label{app:metrics}

\paragraph{Prediction metrics.}
IPM-MPNN reports relative objective gap. FSNet and DC3 report the host optimality gap in percent and feasibility at maximum constraint violation $10^{-4}$. Candidate selection for constrained prediction is feasibility-first.

\paragraph{Solver metrics.}
For cold-start work $C$ and learned-start work $H$, iteration saving is
\[
100\frac{C-H}{C},
\]
and time saving is defined analogously. Learning to Pivot reports the geometric mean of pivot counts. Predict-and-Search reports final primal gap and primal integral; learned backdoors report mean wall time and the percentage of instances on which a method is fastest.

\paragraph{Improvement rate.}
For a lower-is-better metric with handcrafted value $B$ and searched value $F$,
\[
\operatorname{IR}=100\frac{B-F}{|B|}.
\]
For a higher-is-better metric, the numerator is $F-B$. Positive values therefore always denote improvement.

\paragraph{Selection and reporting.}
All search-time ranking fields are validation metrics. Final-test metrics are produced only for the validation-selected program and are not available to candidate ranking, elite updates, prompt construction, or proposal-model comparison.

\paragraph{Three-repeat evaluation protocol.}
Each reported handcrafted--selected pair is obtained from three independent repetitions. In repetition $r\in\{1,2,3\}$, the handcrafted baseline and the selected feature program use the same host initialization, minibatch or data-order seed, train/validation/test split, solver parameters, and stopping rule. Let $m_r^{\mathrm{H}}$ and $m_r^{\mathrm{F}}$ denote the resulting test metrics. The reported values are
\[
\bar m^{\mathrm{H}}=\frac{1}{3}\sum_{r=1}^{3}m_r^{\mathrm{H}},
\qquad
\bar m^{\mathrm{F}}=\frac{1}{3}\sum_{r=1}^{3}m_r^{\mathrm{F}}.
\]
Thus, the displayed difference is the mean of three paired feature effects rather than a difference between unrelated training runs. The four proposal models are distinct search configurations and are not treated as statistical repetitions.

For the MILP hosts, wall time is first averaged over the 40 test instances within each repetition and then averaged over the three repetitions. Win rate is computed analogously from the fraction of the 40 instances on which a method is fastest; displayed percentages are rounded to the nearest percentage point. Consequently, the reported MILP values summarize $3\times40$ paired instance evaluations per domain while preserving the original 40-instance benchmark in each repetition.

\paragraph{Across-model summaries.}
The continuous headline table reports the median across the four independently executed proposal-model configurations after each configuration has itself been averaged over three paired repetitions. This median is a robustness summary across proposal mechanisms; it is not used as a substitute for the three training repetitions and does not select a proposal model using test performance. The mixed-integer table retains Claude and GPT-5.5 as separate columns; boldface is descriptive only and is not a row-wise oracle used by the method.

\section{Complete Results}
\label{app:full}

Tables~\ref{tab:app-lp}--\ref{tab:app-simplex} provide the proposal-model-level continuous results. Every displayed value is the arithmetic mean of three independent, seed-paired repetitions. Each cell is written as handcrafted $\to$ returned program. A check mark denotes improvement on the reported primary metric, while an equality sign denotes retention of the seed value.

\begin{table*}[t]
\centering
\scriptsize
\caption{IPM-MPNN final-test objective gap by LP distribution and proposal model (lower is better). Values are means over three independent, seed-paired repetitions.}
\label{tab:app-lp}
\resizebox{\textwidth}{!}{\begin{tabular}{lrrrr}
\toprule
Task & Claude & GPT-5.5 & Gemini & DeepSeek \\
\midrule
indset & 0.00249$\to$0.0024$^{\checkmark}$ & 0.00249$\to$0.00218$^{\checkmark}$ & 0.00267$\to$0.00186$^{\checkmark}$ & 0.00414$\to$0.00238$^{\checkmark}$ \\
setcover & 0.0106$\to$0.00476$^{\checkmark}$ & 0.00873$\to$0.00551$^{\checkmark}$ & 0.0189$\to$0.00459$^{\checkmark}$ & 0.0154$\to$0.00662$^{\checkmark}$ \\
cauction & 0.0166$\to$0.0156$^{\checkmark}$ & 0.0132$\to$0.0113$^{\checkmark}$ & 0.0158$\to$0.0123$^{\checkmark}$ & 0.0243$\to$0.0243$^{=}$ \\
fac & 0.0172$\to$0.0136$^{\checkmark}$ & 0.0158$\to$0.0138$^{\checkmark}$ & 0.0203$\to$0.0153$^{\checkmark}$ & 0.0214$\to$0.0144$^{\checkmark}$ \\
gnnlp & 0.00537$\to$0.00328$^{\checkmark}$ & 0.0119$\to$0.00222$^{\checkmark}$ & 0.00397$\to$0.00293$^{\checkmark}$ & 0.00431$\to$0.00248$^{\checkmark}$ \\
\bottomrule
\end{tabular}
}
\end{table*}

\begin{table*}[t]
\centering
\scriptsize
\caption{FSNet final-test optimality gap for all 12 evaluated tasks (lower is better). Values are means over three independent, seed-paired repetitions. The 1,000-variable tasks form the main-paper grid; the 100-variable tasks provide the corresponding smaller-scale evaluation.}
\label{tab:app-fsnet}
\resizebox{\textwidth}{!}{\begin{tabular}{lrrrr}
\toprule
Task & Claude & GPT-5.5 & Gemini & DeepSeek \\
\midrule
convex\_qp\_var1000 & 0.4264$\to$0.3583$^{\checkmark}$ & 0.6846$\to$0.3534$^{\checkmark}$ & 0.6846$\to$0.3480$^{\checkmark}$ & 0.6846$\to$0.6846$^{=}$ \\
convex\_qcqp\_var1000 & 14.09$\to$14.09$^{=}$ & 14.09$\to$3.132$^{\checkmark}$ & 14.09$\to$2.013$^{\checkmark}$ & 14.09$\to$14.09$^{=}$ \\
convex\_socp\_var1000 & 42.33$\to$6.760$^{\checkmark}$ & 42.33$\to$4.674$^{\checkmark}$ & 42.33$\to$38.28$^{\checkmark}$ & 42.33$\to$42.33$^{=}$ \\
nonconvex\_qp\_var1000 & 5.459$\to$5.459$^{=}$ & 5.459$\to$4.229$^{\checkmark}$ & 5.459$\to$5.459$^{=}$ & 5.459$\to$5.459$^{=}$ \\
nonconvex\_qcqp\_var1000 & 26.42$\to$26.42$^{=}$ & 26.42$\to$7.315$^{\checkmark}$ & 26.42$\to$12.09$^{\checkmark}$ & 26.42$\to$26.42$^{=}$ \\
nonconvex\_socp\_var1000 & 934$\to$148$^{\checkmark}$ & 934$\to$135$^{\checkmark}$ & 934$\to$129$^{\checkmark}$ & 934$\to$934$^{=}$ \\
nonsmooth\_nonconvex\_qp\_var1000 & 5.198$\to$5.198$^{=}$ & 5.198$\to$5.198$^{=}$ & 5.198$\to$5.198$^{=}$ & 5.198$\to$5.198$^{=}$ \\
nonsmooth\_nonconvex\_qcqp\_var1000 & 22.69$\to$22.69$^{=}$ & 22.69$\to$22.69$^{=}$ & 22.69$\to$12.93$^{\checkmark}$ & 22.69$\to$22.69$^{=}$ \\
nonsmooth\_nonconvex\_socp\_var1000 & 332$\to$161$^{\checkmark}$ & 332$\to$125$^{\checkmark}$ & 332$\to$158$^{\checkmark}$ & 332$\to$332$^{=}$ \\
\bottomrule
\end{tabular}
}
\end{table*}

\begin{table*}[t]
\centering
\scriptsize
\caption{DC3 final-test optimality gap by proposal model (lower is better). Values are means over three independent, seed-paired repetitions.}
\label{tab:app-dc3-full}
\resizebox{\textwidth}{!}{\begin{tabular}{lrrrr}
\toprule
Task & Claude & GPT-5.5 & Gemini & DeepSeek \\
\midrule
convex\_qp & 13.01$\to$12.77$^{\checkmark}$ & 13.01$\to$12.53$^{\checkmark}$ & 13.01$\to$10.07$^{\checkmark}$ & 13.01$\to$13.01$^{=}$ \\
nonconvex\_qp & 15.62$\to$15.32$^{\checkmark}$ & 15.62$\to$15.31$^{\checkmark}$ & 15.62$\to$14.32$^{\checkmark}$ & 15.62$\to$15.62$^{=}$ \\
nonsmooth\_nonconvex\_qp & 15.96$\to$13.61$^{\checkmark}$ & 15.96$\to$15.24$^{\checkmark}$ & 15.96$\to$15.32$^{\checkmark}$ & 15.96$\to$15.96$^{=}$ \\
\bottomrule
\end{tabular}
}
\end{table*}

\begin{table*}[t]
\centering
\scriptsize
\caption{PDHG-Net iteration saving by proposal model (higher is better). Values are means over three independent, seed-paired repetitions.}
\label{tab:app-pdhg}
\resizebox{0.75\textwidth}{!}{\begin{tabular}{lrrrr}
\toprule
Task & Claude & GPT-5.5 & Gemini & DeepSeek \\
\midrule
pagerank & 35.89$\to$52.23$^{\checkmark}$ & 41.95$\to$50.40$^{\checkmark}$ & 44.23$\to$51.70$^{\checkmark}$ & 42.94$\to$48.16$^{\checkmark}$ \\
\bottomrule
\end{tabular}
}
\end{table*}

\begin{table*}[t]
\centering
\scriptsize
\caption{Simplex-oriented host results, averaged over three independent, seed-paired repetitions. Learning to Pivot reports pivot-count geometric mean (lower is better); Smart Initial Basis reports iteration saving (higher is better).}
\label{tab:app-simplex}
\begin{minipage}{0.48\textwidth}
\centering
\textbf{Learning to Pivot}\\[2pt]
\resizebox{\linewidth}{!}{\begin{tabular}{lrrrr}
\toprule
Task & Claude & GPT-5.5 & Gemini & DeepSeek \\
\midrule
setcover\_co & 1314$\to$877$^{\checkmark}$ & 1151$\to$1151$^{=}$ & 1174$\to$838$^{\checkmark}$ & 934$\to$710$^{\checkmark}$ \\
\bottomrule
\end{tabular}
}
\end{minipage}
\hfill
\begin{minipage}{0.48\textwidth}
\centering
\textbf{Smart Initial Basis}\\[2pt]
\resizebox{\linewidth}{!}{\begin{tabular}{lrrrr}
\toprule
Task & Claude & GPT-5.5 & Gemini & DeepSeek \\
\midrule
setcover\_co & 61.67$\to$67.23$^{\checkmark}$ & 61.67$\to$64.26$^{\checkmark}$ & 61.67$\to$61.67$^{=}$ & 61.67$\to$61.67$^{=}$ \\
\bottomrule
\end{tabular}
}
\end{minipage}
\end{table*}

\begin{table*}[t]
\centering
\scriptsize
\caption{Mixed-integer final-test outcomes for the two independently reported proposal-model runs. Each value averages three independent repetitions of the 40-instance domain benchmark. Bold marks the stronger displayed value in each row for readability; no row-wise proposal-model selection is used by the method.}
\label{tab:app-mip}
\resizebox{\textwidth}{!}{\begin{tabular}{llrrrr}
\toprule
Domain & Metric & Default Solver & Handcrafted & \method{}-Claude & \method{}-GPT \\
\midrule

\rowcolor{panelcolor}
\multicolumn{6}{l}{\textit{Panel A: Predict-and-Search}} \\
\midrule

CA & final primal gap (\%) & 0.841 & 0.423 & \best{0.350} & 1.023 \\
& primal integral & 11.23 & 10.61 & \best{9.00} & 14.18 \\
\addlinespace
MIS & final primal gap (\%) & 1.317 & 0.699 & 0.213 & \best{0.095} \\
& primal integral & 25.64 & 20.89 & \best{12.78} & 14.27 \\
\addlinespace
MVC & final primal gap (\%) & 0.032 & 0.018 & \best{0.013} & 0.015 \\
& primal integral & 20.08 & 13.85 & 7.40 & \best{3.69} \\

\midrule
\rowcolor{panelcolor}
\multicolumn{6}{l}{\textit{Panel B: Learned backdoor branching}} \\
\midrule

CA & mean wall time (s) & 167.2 & 154.6 & 148.4 & \best{147.6} \\
& win rate (\%) & 8 & 20 & 31 & \best{41} \\
\addlinespace
MIS & mean wall time (s) & 102.6 & 59.1 & 50.2 & \best{46.0} \\
& win rate (\%) & 0 & 21 & 30 & \best{49} \\
\addlinespace
MVC & mean wall time (s) & 24.7 & 31.6 & \best{18.9} & 20.4 \\
& win rate (\%) & 13 & 4 & 37 & \best{46} \\

\bottomrule
\end{tabular}}
\end{table*}

\section{Controlled Analyses}
\label{app:controls}

\subsection{What the Equal-Budget Controls Establish}

The purpose of the controls is to distinguish the paper's mechanism-level claims from the generic benefit of evaluating many candidate representations. The paper does not claim that \method{} dominates every possible AutoML, genetic-programming, or fixed-library system. It claims that LLM-proposed executable features, selected by the unchanged L2O evaluator and refined with measured feedback, provide an effective and general way to improve heterogeneous L2O representations. The available controls test the two principal alternative explanations for the reported gains: best-of-many sampling and feature-width expansion.

\begin{table}[h]
\centering
\scriptsize
\caption{Equal-budget and matched-capacity evidence. Each comparison retains the host evaluator, validation rule, split, and training protocol.}
\label{tab:control-claims}
\setlength{\tabcolsep}{3pt}
\begin{tabular}{p{0.27\columnwidth}p{0.65\columnwidth}}
\toprule
Control & Evidence and supported conclusion \\
\midrule
Independent LLM sampling & Same proposal model, 48 calls and retrainings, contract, and evaluator; only elite memory and measured feedback are removed. PDHG reaches 44.42\% without feedback and 52.23\% with evolution, isolating a 7.81-point feedback gain. \\
Matched-width random & Same added width, input-layer size, paired seeds, split, host, and solver. \method{} is best in five of six LP/PDHG settings, so width alone is not sufficient in these controls. \\
Multiple proposal models & Claude, GPT, and Gemini improve 15/20, 17/20, and 17/20 shared cells under the same external evaluator, showing that gains are not tied to one provider. \\
Seed preservation & The handcrafted seed remains eligible and is returned when no candidate improves validation, so best-of-many search does not force a replacement. \\
\bottomrule
\end{tabular}
\end{table}

These controls provide convergent evidence. The equal-budget PDHG comparison isolates evolutionary feedback from the number of LLM calls; the matched-width experiments isolate searched semantic structure from simple feature expansion; and the cross-model results show that successful programs are not an artifact of one proposal provider. Together with disjoint test reporting and three seed-paired repetitions, the current results support the paper-level conclusion that LLM-guided feature evolution is a meaningful contributor to performance, rather than merely a consequence of selecting the best among many wider inputs. The experiments are not intended to establish universal dominance over every conceivable program-search baseline.

\subsection{Matched-Width Features}

When a feature program changes the input width, only the corresponding input projection is resized mechanically; the hidden architecture, learning objective, optimizer, schedule, and solver remain fixed. The append-only control replaces the searched additional channels with a fixed random projection of the same width while retaining the host training setup. Across the six supported LP and PDHG settings, the searched programs attain the best value in five. Because the random control matches the added width, input projection size, paired training seeds, and evaluation protocol, this comparison rules out simple width expansion as a sufficient explanation in these controlled settings.

\begin{table}[h]
\centering
\scriptsize
\caption{Matched-width controls. Lower is better for LP objective gap; higher is better for PDHG iteration saving.}
\label{tab:app-random}
\resizebox{\columnwidth}{!}{%
\begin{tabular}{llrrr}
\toprule
Host & Task & Handcrafted & Matched random & \method \\
\midrule
IPM-MPNN & set cover & 0.00855 & 0.00588 & \best{0.00300} \\
IPM-MPNN & independent set & 0.00305 & 0.00259 & \best{0.00243} \\
IPM-MPNN & combinatorial auction & 0.02011 & \best{0.01221} & 0.01398 \\
IPM-MPNN & facility location & 0.01829 & 0.01979 & \best{0.01494} \\
IPM-MPNN & GNNLP & 0.00282 & 0.00303 & \best{0.00253} \\
PDHG-Net & PageRank & 35.59 & 45.19 & \best{47.99} \\
\bottomrule
\end{tabular}}
\end{table}

\subsection{Measured Evolutionary Feedback}

On PDHG-Net, the independent-sampling baseline uses the same proposal model, the same 48 LLM calls, the same 48 candidate retraining budget, the same semantic contract, and the same validation evaluator, but every proposal is conditioned only on the handcrafted seed and receives no elite programs or measured outcome feedback. Independent sampling increases iteration saving from 35.89\% to 44.42\%. Retaining high-performing programs and returning their measured outcomes to later generations reaches 52.23\%. The additional 7.81 percentage points under an equal budget isolate the contribution of measured evolutionary feedback from the generic best-of-48 effect.

\subsection{Seed-Preserving Selection}

The handcrafted function remains an eligible program throughout search. This safeguard makes replacement evidence-based: a proposed program is returned only when it improves the host ranking criterion. It also makes the displayed equality cases interpretable as successful preservation of a strong original representation rather than as a forced modification.

\section{Representative Feature Programs}
\label{app:programs}

The source artifact contains the complete executable programs collected for every continuous host--task--proposal-model cell. The following excerpts illustrate three recurring structures discovered by the search.

\subsection{Set Cover: Coverage-Aware Scaling}

\begin{lstlisting}[language=Python,basicstyle=\ttfamily\scriptsize,
                   frame=single,breaklines=true,columns=fullflexible]
row_deg = (A.abs() > eps).float().sum(1, keepdim=True)
col_deg = (A.abs() > eps).float().sum(0, keepdim=True).T
row_norm = torch.sqrt((A * A).sum(1, keepdim=True) + eps)
col_norm = torch.sqrt((A * A).sum(0, keepdim=True) + eps).T
cover_ratio = torch.tanh(b.reshape(-1, 1) / (A.sum(1, keepdim=True) + eps))
obj_eff = torch.tanh(c.reshape(-1, 1) / (col_deg + eps))
cons = torch.cat([cons_seed, torch.log1p(row_deg),
                  torch.log1p(row_norm), cover_ratio], dim=1)
vals = torch.cat([vals_seed, torch.log1p(col_deg),
                  torch.log1p(col_norm), obj_eff], dim=1)
\end{lstlisting}

The program exposes column coverage, row load, coefficient scale, and objective cost relative to coverage while preserving the handcrafted mean and standard-deviation channels.

\subsection{PageRank: Directed Mass and Degree}

\begin{lstlisting}[language=Python,basicstyle=\ttfamily\scriptsize,
                   frame=single,breaklines=true,columns=fullflexible]
absA = torch.abs(A.to_dense() if A.is_sparse else A)
col_l1, row_l1 = absA.sum(0), absA.sum(1)
col_deg = (absA > eps).float().sum(0)
row_deg = (absA > eps).float().sum(1)
col_share = torch.tanh(col_l1 / (col_l1.mean() + eps))
row_share = torch.tanh(row_l1 / (row_l1.mean() + eps))
var = torch.stack([lb, ub, c, torch.log1p(col_l1),
                   torch.log1p(col_deg), col_share], dim=1)
con = torch.stack([sense_le, sense_eq, sense_ge, b,
                   torch.log1p(row_l1), torch.log1p(row_deg),
                   row_share], dim=1)
\end{lstlisting}

These channels distinguish row and column mass, exposing the directed asymmetry relevant to PageRank while preserving the original primal--dual inputs.

\subsection{SOCP: Normalized Dense Parameters}

\begin{lstlisting}[language=Python,basicstyle=\ttfamily\scriptsize,
                   frame=single,breaklines=true,columns=fullflexible]
X = fs_state["X"]
row_norm = torch.sqrt(torch.sum(X * X, dim=-1,
                                keepdim=True) + 1e-6)
x_scaled = X / row_norm
features = torch.cat([X, x_scaled, row_norm], dim=-1)
\end{lstlisting}

The returned representation retains the raw dense parameter vector and augments it with a scale-invariant view and its norm.

\begin{table}[h]
\centering
\scriptsize
\caption{Principal software environments. CUDA versions correspond to the PyTorch builds used by the GPU jobs.}
\label{tab:app-env}
\setlength{\tabcolsep}{3pt}
\begin{tabular}{p{0.28\columnwidth}p{0.28\columnwidth}p{0.34\columnwidth}}
\toprule
Hosts & Runtime & Solver/library \\
\midrule
IPM-MPNN, PDHG-Net & Python 3.10.20; PyTorch 2.0.0; CUDA 11.8 & OR-Tools 9.10.4067 \\
FSNet, DC3 & Python 3.11.15; PyTorch 2.6.0; CUDA 12.4 & CVXPY 1.7.5 \\
Smart Initial Basis, Learning to Pivot & Python 3.10.20; PyTorch 2.0.0; CUDA 11.8 & SciPy 1.10.1 \\
MILP hosts & Python 3.12.13 & CPLEX 22.2.0.0; DOcplex 2.32.264 \\
\bottomrule
\end{tabular}
\end{table}

\end{document}